\documentclass[11pt]{article}
\PassOptionsToPackage{dvipsnames,table}{xcolor}
\usepackage{latex/acl}

\usepackage{times}
\usepackage{latexsym}
\usepackage{booktabs}
\usepackage{graphicx}
\usepackage{graphics}
\usepackage{amsmath}
\usepackage{multirow}
\usepackage{tabu}
\usepackage{subcaption}
\usepackage[T1]{fontenc}
\usepackage{tabularx}
\usepackage{soul}

\definecolor{one}{rgb}{0.33999999999999997, 0.8287999999999999, 0.86}
\definecolor{two}{rgb}{0.5688000000000001, 0.86, 0.33999999999999997}

\DeclareRobustCommand{\hlone}[1]{{\sethlcolor{one!40}\hl{#1}}}
\DeclareRobustCommand{\hltwo}[1]{{\sethlcolor{two!40}\hl{#1}}}

\usepackage[utf8]{inputenc}

\usepackage{microtype}

\usepackage{inconsolata}

\usepackage{tcolorbox}
\tcbuselibrary{breakable}

\usepackage{enumitem}
\setlist{leftmargin=*,topsep=0pt,itemsep=0pt,partopsep=1ex,parsep=1ex}

\usepackage{todonotes}
\makeatletter
\newcommand*\iftodonotes{\if@todonotes@disabled\expandafter\@secondoftwo\else\expandafter\@firstoftwo\fi}
\makeatother

\definecolor{edolime}{rgb}{0.9,1,0.3}

\title{Cross-Lingual and Cross-Cultural Variation in Image Descriptions}

\author{Uri Berger \\
  Hebrew University of Jerusalem \\University of Melbourne \\
  \texttt{uri.berger2@mail.huji.ac.il} \\\And
  Edoardo M. Ponti \\
  University of Edinburgh \\
  University of Cambridge \\
  \texttt{eponti@ed.ac.uk} \\}

\begin{document}
\maketitle
\begin{abstract}
Do speakers of different languages talk differently about what they see? Behavioural and cognitive studies report cultural effects on perception; however, these are mostly limited in scope and hard to replicate. In this work, we conduct the first large-scale empirical study of cross-lingual variation in image descriptions. Using a multimodal dataset with 31 languages and images from diverse locations, we develop a method to accurately identify entities mentioned in captions and present in the images, then measure how they vary across languages.
Our analysis reveals that pairs of languages that are geographically or genetically closer tend to mention the same entities more frequently.
We also identify entity categories whose saliency is universally high (such as animate beings), low (clothing accessories) or displaying high variance across languages (landscape).
In a case study, we measure the differences in a specific language pair (e.g., Japanese mentions clothing far more frequently than English).
Furthermore, our method corroborates previous small-scale studies, including 1) \citet{rosch1976basic}'s theory of basic-level categories, demonstrating a preference for entities that are neither too generic nor too specific, and 2) \citet{miyamoto2006culture}'s hypothesis that environments afford patterns of perception, such as entity counts. 
Overall, our work reveals the presence of both universal and culture-specific patterns in entity mentions.\footnote{We provide our code at\\ \href{https://github.com/uriberger/cross_lingual_diff_in_descs}{github.com/uriberger/cross\_lingual\_diff\_in\_descs}
and a web interface for our dataset at \href{https://tinyurl.com/5ekabrb5}{tinyurl.com/5ekabrb5}.
}
\end{abstract}

\section{Introduction}
\begin{figure} [tb]
    \includegraphics[width=\columnwidth,trim={0 4cm 0 0},clip]{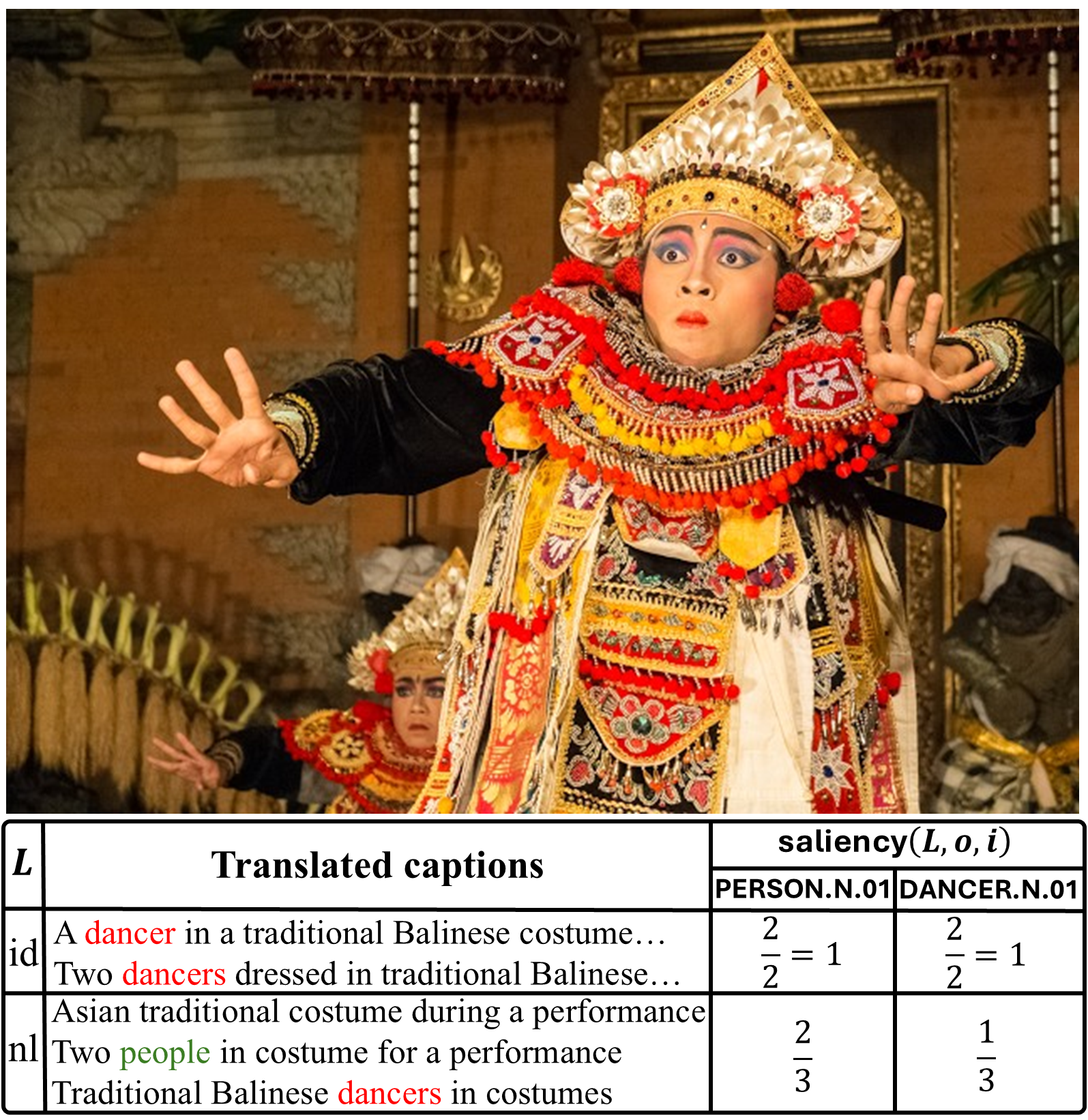}
    
    \resizebox{\columnwidth}{!}{%
    \begin{tabular}{l|l|c|c}
        \toprule
        \multirow{2}{*}{} & \multirow{2}{*}{Translated captions} & \multicolumn{2}{c}{Saliency} \\
        \cline{3-4}
        & & \textsc{person.n.01} & \textsc{dancer.n.01} \\
        \midrule
        \multirow{2}{*}{id} & A \hlone{dancer} in a traditional Balinese costume\dots & \cellcolor{red!30} & \cellcolor{red!30} \\
        & Two \hlone{dancers} dressed in traditional Balinese\dots & {\LARGE \multirow{-2}{*}{\cellcolor{red!30} $\frac{2}{2} = 1$}} & {\LARGE \multirow{-2}{*}{\cellcolor{red!30} $\frac{2}{2} = 1$}} \\
        \hline
        \multirow{3}{*}{nl} & Asian traditional costume during a performance & \cellcolor{red!20} & \cellcolor{red!10}   \\
        & Two \hltwo{people} in costume for a performance & \cellcolor{red!20}  & \cellcolor{red!10} \\
        & Traditional Balinese \hlone{dancers} in costumes & {\LARGE \multirow{-3}{*}{\cellcolor{red!20} $\frac{2}{3}$}} & {\LARGE \multirow{-3}{*}{\cellcolor{red!10} $\frac{1}{3}$}} \\
        \bottomrule
    \end{tabular}%
    }
    \caption{A photo taken in an Indonesian-speaking area, corresponding captions in Indonesian (id) and Dutch (nl) translated to English, and saliency for the {\textsc{person.n.01}} and \textsc{dancer.n.01} synsets. Saliency is measured as the proportion of captions referring to the synset or any of its descendants (e.g., \textsc{dancer.n.01} is a descendant of \textsc{person.n.01}).}
    \label{fig:image_example}
\end{figure}

Do speakers of different languages talk differently about what they see? This question is important for both machine learning and cognitive science. From a computational perspective, the answer can guide the creation of more diverse vision-and-language datasets \citep{ye2023cultural}. From a cognitive science perspective, it can help us study cultural effects on perception and language.

To investigate their cross-lingual and cross-cultural variation, we can compare the descriptions of the same images by different communities of speakers.
In this setting,
differences in semantic content must stem from linguistic rather than perceptual differences.\footnote{It must be noted that \citet{mitterer2009influence} provide evidence of environmental effects on visual perception; however, these effects are limited to low-level details, such as similar colour discrimination.} Crucially, describing the same image takes into account cultural effects (e.g., the saliency of an entity) while excluding environmental ones (e.g., the presence of an entity) \citep{liu-etal-2021-visually,hershcovich-etal-2022-challenges}.

Previous work investigated which cultures draw more attention to the foreground or the context of a visual scene~\cite{miyamoto2006culture}, and how often each culture uses negation or certain bigrams at the start of the descriptions~\cite{van-miltenburg-etal-2017-cross}.
Nevertheless, these studies were conducted on a small scale, either with artificial images or focused on just a few related languages. 

To induce more solid conclusions, a large dataset of image descriptions in a variety of typologically and areally diverse languages is needed \citep{typologynlp}. We use Crossmodal3600~\citep[XM3600,][]{thapliyal-etal-2022-crossmodal}, which is both relatively large in scale (3600 images) compared with previous controlled studies and more linguistically diverse (36 languages) compared with any previous image--description dataset.

We develop automatic tools for accurately identifying phrases mentioning entities in image descriptions and map these phrases into WordNet~\cite{miller1995wordnet} synsets, which function as a proxy for entity categories. 
We further annotate images with the categories they contain, independent of the descriptions.
Finally, we study how entity mentions vary across different languages  (see Figure~\ref{fig:image_example} for an example).

Using this annotation as quantitative evidence, we are able to address the following questions:

\begin{tcolorbox}{%
\begin{enumerate}
    \item Do \textbf{geographically, genetically, or typologically similar} languages share the same saliency of entities?
    \item Which entities are \textbf{universally} (non-)sa\-lient and which ones display the \textbf{largest variance} across languages?
    \item What is the semantic \textbf{granularity} of entities mentioned in different languages?
    \item Is the \textbf{number} of entities mentioned affected by the familiarity or the complexity of the images?
\end{enumerate}
}%
\end{tcolorbox}

Our method can be used both as an exploration and a verification tool for cultural effects on object saliency.
For the former, we present new findings (Sections~\ref{sec:typology_saliency_corr}, \ref{sec:cross_lingual_var}, and \ref{sec:en_ja_saliency}). Future work may propose small-scale, controlled studies to verify these findings.

For the latter, we demonstrate how our method can validate previous findings from cognitive studies.
First, annotators typically select object categories that fall in the middle of the hierarchy, with more general (lower) and more specific (higher) categories being less frequent, supporting \citet{rosch1976basic}'s hypothesis of basic-level categories (Section~\ref{sec:granularity}).
Second, we put to the test \citet{miyamoto2006culture}'s claim that images taken in East Asian locations tend to include more entities compared to Western locations (Section~\ref{sec:object_num}).

To summarise, our contributions are as follows. First, we present new insights into cross-lingual variation in entity saliency and confirm previous findings on a wider and more diverse set of languages. Second, we release our entity mention dataset to support future research on cross-lingual and cross-cultural variation of image descriptions.

\section{Background and Related Work}
Our study falls under the umbrella of works that study cultural effects on perception and language. Early studies relied on small-scale, controlled experiments, while the recent rise in the amount of publicly available visual data with human annotations has prompted large-scale and real-world studies harnessing computational methods to investigate these datasets.

\subsection{Small-Scale Controlled Studies}

Numerous studies detail experiments involving human participants, who are grouped according to their cultural backgrounds and tasked with completing a visual assignment. Subsequently, the outcomes are examined and compared across different groups.
Relevant visual tasks include, among many others, colour discrimination~\cite{winawer2007russian, mitterer2009influence}, identification of a change in an observed scene~\cite{miyamoto2006culture}, visual illusions~\cite{segall1963cultural, caparos2012exposure, linnell2018urban}, and reading facial expressions~\cite{jack2009cultural}. 

Most relevant to the present work, \citet{masuda2001attending} study differences in image descriptions across cultures. They display animated clips of underwater scenes to Japanese and Americans and subsequently collect their description of the scenes. The Japanese participants provided more statements about contextual information and relationships, rather than the foreground. \citet{senzaki2014perception} replicate this result for the Canadian and Japanese, while \citet{senzaki2016communication} demonstrate that the effect is more pronounced for adult participants compared to children.

\citet{miyamoto2006culture} put forth the hypothesis that these differences are driven by the typical environments encountered by speakers of different cultures. In particular, they show that photos taken in Japan contain more entities according to both human annotators and automated counts, and that American participants attend more on contextual information when primed with Japanese photos than with American photos.

\subsection{Large-Scale Studies}
Several previous studies have recognised the advantage of utilising existing public image--caption datasets as a test bed for large-scale experiments of cultural effects on perception and language.

\Citet{van-miltenburg-etal-2017-cross} compare descriptions generated by English, German, and Dutch annotators on images taken from Flickr30k~\cite{young2014image}. They study lexical differences such as the use of negation and the degree of specificity in annotations across languages. 
Their approach diverges from ours in several key respects. First, we focus on XM3600, which contains culturally diverse images and covers a wider range of languages. Crucially, we systematically study patterns of entity saliency by developing automated tools for entity mention identification; their method instead involves manually inspecting specific images.

\citet{ye2023cultural} focus on cross-lingual differences in descriptions of 7 languages from XM3600.
Their motivation is computational, aiming to demonstrate that annotators from different languages highlight distinct entities, leading to a more comprehensive coverage of entities in multilingual image-caption datasets compared to monolingual ones. This, in turn, offers improved training data for vision-and-language foundation models.
This echoes similar considerations from \citet{liu-etal-2021-visually}, who created the MaRVL dataset by sourcing images from 5 diverse cultures representing culturally salient entities.
In contrast, our motivation is cognitive; we recognise the existence of such cross-lingual variation in image annotation and leverage it as a framework to investigate questions about cross-cultural variation.

\subsection{WordNet}

WordNet~\cite{miller1995wordnet} is a large database of syn\-sets arranged into a tree structure of hyponyms and hypernyms (among other lexical relations). Each synset contains all the English words that are synonyms according to one of their senses. For instance, the synset \textsc{bank.n.01} represents the \emph{river bank} meaning of the word \textit{bank} and is the direct child of the \textsc{slope.n.01} synset, while the synset \textsc{bank.n.02} represents the \emph{financial institution bank} meaning and is the direct child of the \textsc{financial\_institution.n.01} synset.

\section{Methods}
To identify the entities mentioned in a caption, we use WordNet's synsets as a proxy for entity categories. We translate all captions to English (Section~\ref{sec:caption_translation}), pre-define a target list of synsets (Section~\ref{sec:synset_selection}), extract noun phrases that correspond to one of these synsets (Section~\ref{sec:synset_extraction}) and filter the resulting synset lists to remove synsets corresponding to entities that are not in the image (Section~\ref{sec:synset_filtering}).

\subsection{Captions Translation}
\label{sec:caption_translation}

We first translate captions from all languages into English.
Translation is a necessity if we are to consider a wide variety of languages since the tools required to automatically process the captions (e.g., WordNet and part-of-speech taggers) are only available in a few languages. Nevertheless, the translation process presents certain benefits, notably the elimination of metaphorical expressions containing entities in non-English languages, and thus a possible confounding factor.

Translation quality plays a critical role in our experiments, as incorrect translations can impact our conclusions.
We therefore use the Google Translation API\footnote{\href{https://cloud.google.com/translate}{https://cloud.google.com/translate}}, which demonstrates robust performance in 31 out of the 36 languages in XM3600 according to several studies~\cite{jiao2023chatgpt, enis2024llm}.\footnote{Arabic (ar), Chinese-Simplified (zh), Croatian (hr), Czech (cs), Danish (da), Dutch (nl), English (en), Filipino (fil), Finnish (fi), French (fr), German (de), Greek (el), Hebrew (he), Hindi (hi), Hungarian (hu), Indonesian (id), Italian (it), Japanese (ja), Korean (ko), Norwegian (no), Persian (fa), Polish (pl), Portuguese (pt), Romanian (ro), Russian (ru), Spanish (es), Swedish (sv), Thai (th), Turkish (tr), Ukrainian (uk), Vietnamese (vi).}
We exclude the remaining 5 low-resource languages\footnote{Bengali (bn), Maori (mi), Cusco Quechua (quz), Swahili (sw), Telugu (te).}
from our experiments, as the Google Translation API is rated poorly by native speakers of these languages~\cite{benjamin2019teach}.

\subsection{Synset Selection} \label{sec:synset_selection}

We select a subset of synsets to be identified as entities based on their use in XM3600 and their relevance to our study. We start by mapping all noun phrases in (translated) XM3600 captions to WordNet synsets.
Afterwards, we select \emph{root synsets}, i.e., synsets that (a) can be visibly identified as present or absent in an image (unlike, for example, \textsc{city.n.01}), (b) have no ancestors in the WordNet hierarchy that meet condition a, and (c) have at least 100 instantiations of the synset or its descendants in the captions, to provide sufficient statistical strength
(the list of root synsets is provided in Appendix~\ref{sec:app_synsets}).
Next, we include \emph{explicit synsets}, i.e., all descendants of the root synsets to which at least 100 noun phrases from XM3600 captions were mapped. Finally, we include all intermediate synsets between a root synset and its descendent explicit synsets (\emph{implicit synsets}). For instance, \textsc{animal.n.01} is a root synset and \textsc{squirrel.n.01} is an explicit synset: we therefore add all the synsets along the path in the tree between the two: \textsc{chordate.n.01}, \textsc{vertebrate.n.01}, \textsc{mammal.n.01}, \textsc{placental.n.01}, \textsc{rodent.n.01}. As a result, we obtain a list of 649 synsets (25 root, 391 explicit, 233 implicit).

\subsection{Synset Extraction} \label{sec:synset_extraction}

We then extract the selected synsets from the captions as follows. 
We classify the words in the caption with off-the-shelf part-of-speech taggers, extract noun phrases, identify WordNet synsets (if any exist) each noun phrase should be mapped to, and resolve ambiguities when a noun phrase may refer to multiple synsets. We now elaborate on each of these steps. Additional details can be found in Appendix~\ref{sec:app_extraction}.

\paragraph{Noun phrase extraction.}
We assign each word in the captions a part-of-speech tag with Stanza~\cite{qi2020stanza}.  Then,
we extract noun phrases by identifying all sequences in the caption that consist of one or more consecutive nouns.

\paragraph{Synset identification.}
We then link each noun phrase to its corresponding WordNet synsets with the NLTK API.\footnote{\href{https://www.nltk.org/howto/wordnet.html}{nltk.org/howto/wordnet.html}} If one of these synsets, or one of its ancestor synsets, is found in our predefined selected synset list (see Section~\ref{sec:synset_selection}), we assign the relevant synset to this phrase. Note that, since WordNet may map phrases to multiple synsets, by the end of this phase each noun phrase may be assigned with zero, one, or more synsets.

\paragraph{Resolving ambiguities in synset mapping.}
If a noun phrase is assigned multiple synsets, we resolve this ambiguity through a pre-trained language model---specifically, BERT large uncased~\cite{devlin2018bert}. Given a caption and candidate synsets $s_1, ..., s_n$ where $n > 1$, we generate $n$ new versions of the caption, where in sentence $c_i$ we replace the phrase in question with a phrase representing synset $s_i$. This representative phrase could be the synset name or an example of the synset. 
BERT then calculates the probability of each synset phrase via masked language modelling. We select the synset with the highest probability.

\subsection{Synset Filtering} \label{sec:synset_filtering}
The list of synsets generated by the aforementioned process may include synsets that correspond to entities not present in the associated image for several reasons. First, a synset might mistakenly appear on the list due to errors of captioning annotators, automatic translation, or our synset extraction method (Section~\ref{sec:synset_extraction}). Second, annotators may mention entities that are not visible in the image. For example, in the caption \emph{A woman standing with her back to the camera}, the synset \textsc{camera.n.01} will be extracted even though no camera is actually depicted in the image.

Since we are interested in patterns of entity mentions for entities actually present in the image, we aim to filter out such synsets. We therefore manually annotate each image in the XM3600 dataset, identifying which of the 25 root synsets have instances in the image. We then refine the synset list extracted using the process in Section~\ref{sec:synset_extraction} by removing any synsets whose corresponding root synset is not represented in the image.

\subsection{Validation}

To validate our synset extraction method, we randomly sample 5 captions per language, for a total of 155 captions, manually annotate the synsets mentioned in the image description, and compute the precision and recall of the synset list extracted using our method.

In this sample, our method extracted 247 synsets, 240 of which matched the ground truth identified during annotation, resulting in a precision of 0.97. We manually annotated 247 synsets, 240 of which were correctly predicted by our method, yielding a recall of 0.97.

\section{Experiments} \label{sec:experiments}
In this section we experiment on our entity mentions corpus, extracted from XM3600 captions.

\subsection{Definitions}
Each example in XM3600 associates an image with multiple captions in each language $l \in \mathcal{L}$: $\mathcal{D}_l = \{i, \{c_j\}_{j=1}^n\}_{i=1}^{3600}$.
For a given entity $o \in \mathcal{O}$ (i.e., WordNet synset) and caption (a sequence of words) $c \in \mathcal{V}^+$, we define the function $f(o, c)$ as returning 1 if $o$ is in the list of entities extracted from $c$, and 0 otherwise.
If a synset is extracted for a given caption, all its ancestors in the WordNet tree are also considered extracted.

\paragraph{Entity saliency.}

We define the saliency of entity $o$ in image $i$ as the fraction of captions of this image for which $o$ is in the list of extracted entities.
Formally, given
captions for $i$ in multiple languages
$\mathcal{L}_1, ..., \mathcal{L}_m$,
\begin{equation*}
    \begin{matrix}
    c^{\mathcal{L}_1}_{1}, & c^{\mathcal{L}_1}_{2}, & ..., & c^{\mathcal{L}_1}_{n_1} \\
    c^{\mathcal{L}_2}_{1}, & c^{\mathcal{L}_2}_{2}, &..., &c^{\mathcal{L}_2}_{n_2} \\
    ..., & ..., & ... & ... \\
    c^{\mathcal{L}_m}_{1}, & c^{\mathcal{L}_m}_{2}, &..., &c^{\mathcal{L}_m}_{n_m}
    \end{matrix}
\end{equation*}
we denote the saliency of $o$ in $i$ for captions in language $l \in \mathcal{L}$ as
\begin{equation} \label{eq:ls_saliency}
    \text{saliency}(l, o, i) = \frac{\sum_{j=1}^{n_l}{f(o, c^{l}_{j})}}{n_l}
\end{equation}
See Figure~\ref{fig:image_example} for an example. Composing the saliency scores across entities, languages, and images, we obtain a tensor of size $M \in [0, 1]^{|\mathcal{L}|, |\mathcal{O}|, |\mathcal{I}|}$. The cell $M_{l, o, i}$ represents the saliency of entity $o$ in language $l$ for image $i$.

\subsection{Geographic and Genetic Effects} \label{sec:typology_saliency_corr}

We first study whether languages that are geographically close, genetically related, or typologically similar, also share similar patterns in terms of entity saliency. To this end, we correlate two distance metrics: typological distance and saliency distance between languages.

\paragraph{Typological distance.}
We use \texttt{lang2vec}~\cite{littell2017uriel} to source pre-computed geographical, genetic and featural distances between languages. Featural distance is the overlap between phonemic inventories.

\paragraph{Saliency distance.}
We define the saliency distance between two languages $(\mathcal{L}_l, \mathcal{L}_k)$ as follows.
Given the tensor of saliency scores $M$ obtained from Equation~\ref{eq:ls_saliency}, 
we measure saliency distance as the Euclidean distance between language-specific scores: $\sqrt{\sum_o^{|\mathcal{O}|} \sum_i^ {|\mathcal{I}|}[(M_{\mathcal{L}_l, o, i}) - (M_{\mathcal{L}_k, o, i})]^2}$

\begin{table}[t]
\centering
\begin{tabular}{crrr}
\toprule
\textbf{Typological Criterion} & $r$ & $p$ & $z$ \\
\midrule
Geographic & 0.34 & 0.04* & 2.12 \\
Genetic & 0.26 & 0.01* & 2.51 \\
Featural & -0.01 & 0.92\phantom{*} & -0.11 \\
\bottomrule
    \end{tabular}
    \caption{Mantel test between the saliency distance matrix (Euclidean distance of saliency scores) and typological distance matrices. * denotes statistical significance.}
    \label{tab:mantel}
\end{table}

\paragraph{Correlation significance test.}
For each distance metric, we construct a distance matrix where the $(l, k)$ cell represents the distance between $\mathcal{L}_l$ and $\mathcal{L}_k$ according to that metric.
We use the Mantel test~\cite{mantel1967detection} to compute the correlation between the saliency distance matrix and each of the typological distance matrices. Results are presented in Table~\ref{tab:mantel}. We find that saliency distance correlates weakly with both geographic and genetic distances based on $r \in [-1, 1]$ scores, but $p$-values are statistically significant. On the other hand, we find no significant correlation with featural distance.

\begin{figure*}[t]
    \centering
    \includegraphics[width=\linewidth]
    {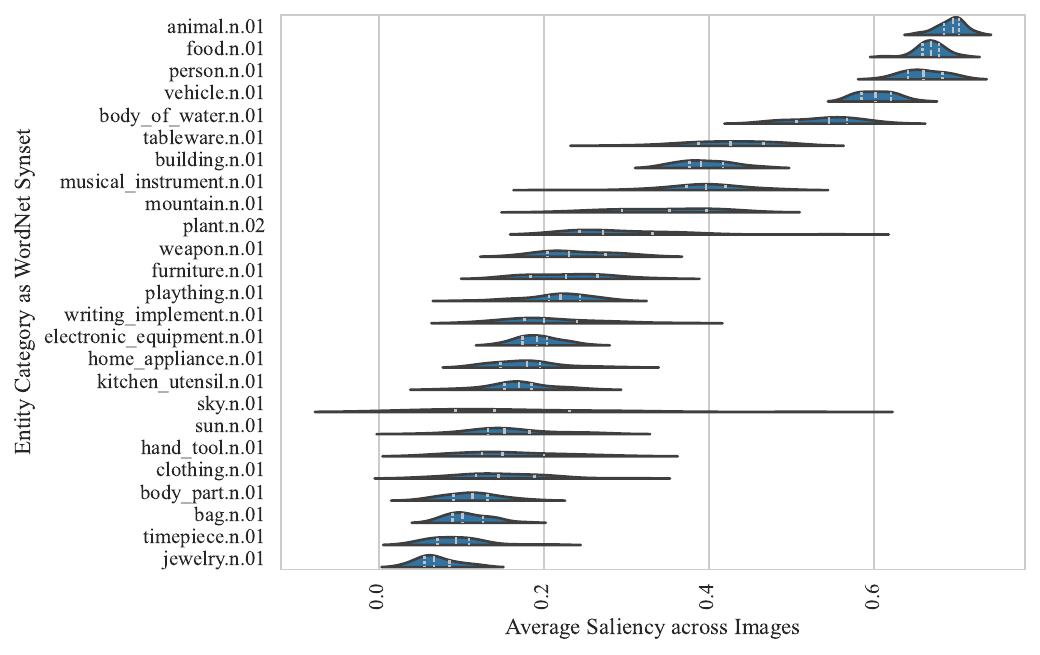}
    \caption{Violin plot of the distribution of saliency scores across languages for each entity category. Saliency scores are averaged across all images containing the entity. Silver lines indicate quartiles.}
    \label{fig:violin_saliency_variation}
\end{figure*}

\subsection{Cross-lingual Variation of Entity Saliency} \label{sec:cross_lingual_var}
Next, we investigate which entity categories exhibit cross-lingual variation in saliency. In other words, which entities are universally salient or non-salient? The saliency of which ones varies the most across languages?

Let $\mathcal{I}_o \subseteq \mathcal{I}$ identify the subset of images where entity $o$ is visible according to our manual image annotation mentioned in Section~\ref{sec:synset_filtering}. Then we calculate the global saliency of an 
entity $o$ and language $l$ as the mean saliency across images $i \in \mathcal{I}_o$.

We plot the resulting global saliency scores in Figure~\ref{fig:violin_saliency_variation}: for each entity category, we report their distribution across languages.
First, we find that certain entities are universally salient: for instance, \textsc{animal.n.01} (0.693), \textsc{food.n.01} (0.669), and \textsc{person.n.01} (0.660). Remarkably, these include the only two entities with animacy \citep{dahl2000animacy}. Conversely, we found that some entities are universally non-salient, e.g., \textsc{bag.n.01} (0.109), \textsc{timepiece.n.01} (0.093), and
\textsc{jewelry.n.01} (0.070). These broadly fall into fashion accessories and apparel.

Finally, Figure~\ref{fig:violin_saliency_variation} lets us identify entities whose global saliency exhibits the highest standard deviation across languages, i.e., \textsc{sky.n.01} (0.109), \textsc{plant.n.02} (0.069), \textsc{mountain.n.01} (0.063).
This suggests that the inclusion of entities commonly located in the background in the descriptions varies between languages, consistent with findings from previous small-scale, controlled studies~\citep[e.g.,][]{miyamoto2006culture}.

\subsection{Case Study: English versus Japanese} \label{sec:en_ja_saliency}
While the results in Section~\ref{sec:cross_lingual_var} inform us of global cross-linguistic tendencies, a finer-grained analysis could try to determine the differences in saliency between specific pairs of languages.
Nevertheless, this raises the concern that the data collected for XM3600 may be insufficient, as it employed only two annotators per language.
Thus, any observed effect would reflect the preferences of those individuals rather than a broader trend.

To mitigate this potential drawback, we focus on a case study for a language pair, namely English--Japanese, for which a significantly larger parallel image captioning dataset is available.
Specifically, the MSCOCO~\cite{lin2014microsoft} dataset comprises a substantial collection of English image captions, encompassing approximately 120K images, each accompanied by an average of five captions. STAIR-captions~\cite{yoshikawa2017stair} is a Japanese version of MSCOCO, wherein Japanese annotators have provided original captions, without resorting to translation, for each image, yielding five captions per image. Overall, about 2100 annotators contributed to the STAIR-captions dataset.

Similarly to Section~\ref{sec:cross_lingual_var}, we define the global saliency for a specific entity category in each language as the mean saliency across all images in MSCOCO in the captions for that language.
For each root synset, we calculate the ratio between the English and Japanese per-language saliency. Table~\ref{tab:en_ja_saliency} plots the top five root synsets in terms of saliency ratio, in both directions.
While we only report results for MSCOCO due to the small sample size of XM3600, we do note that similar rankings were found with XM3600, where the synsets with the highest saliency ratios between Japanese and English were \textsc{clothing.n.01} (3.78) and \textsc{body\_part.n.01} (3.52).
As expected, the concepts where we record the highest differences between Japanese and English are also among those with the least saliency cross-linguistically, as they lie at the bottom of Figure~\ref{fig:violin_saliency_variation}.

\begin{table}[t]
\centering
\begin{tabular}{lc}
\toprule
Synset & Saliency ratio \\
\midrule
& en/ja \\
\midrule
\textsc{writing\_implement.n.01}* & 2.06 \\
\textsc{kitchen\_utensil.n.01}* & 1.43 \\
\textsc{building.n.01}* & 1.42 \\
\textsc{furniture.n.01}* & 1.39 \\
\textsc{body\_of\_water.n.01}* & 1.27 \\
\midrule
& ja/en \\
\midrule
\textsc{clothing.n.01}* & 2.09 \\
\textsc{body\_part.n.01}* & 1.51 \\
\textsc{mountain.n.01}* & 1.42 \\
\textsc{musical\_instrument.n.01} & 1.32 \\
\textsc{bag.n.01}* & 1.18 \\
\bottomrule
    \end{tabular}
    \caption{Top saliency ratios in favor of English (top) and Japanese (bottom). * denotes statistical significance.}
    \label{tab:en_ja_saliency}
\end{table}

\paragraph{Statistical significance.}
To assess statistical significance, we conduct a comparison between the lists of global saliency values associated with each synset in the two languages.
We use the Wilcoxon signed-ranked test~\cite{wilcoxon1992individual} with Bonferroni correction. All entities in Table~\ref{tab:en_ja_saliency} except \textsc{musical\_instrument.n.01} show significance.

\paragraph{Qualitative Analysis: Clothing.}
We now zoom in further onto a specific category, showcasing how our method can serve as an exploration tool for identifying preliminary findings that can later be validated through controlled experiments.
We focus on clothing-related entities, which showed the highest saliency ratio between Japanese and English.
To gain deeper insights into the origin of this disparity, we conduct the statistical significance test for all the explicit synsets that are descendants (see Section~\ref{sec:synset_selection}) of the \textsc{clothing.n.01} synset. We find that only the \textsc{shirt.n.01} synset was significantly more salient in Japanese in both MSCOCO and XM3600.
This indicates that differences in high-level categories may be driven by specific (culturally motivated) entities.

\begin{figure}[t!]
    \centering
    \includegraphics[width=\columnwidth]{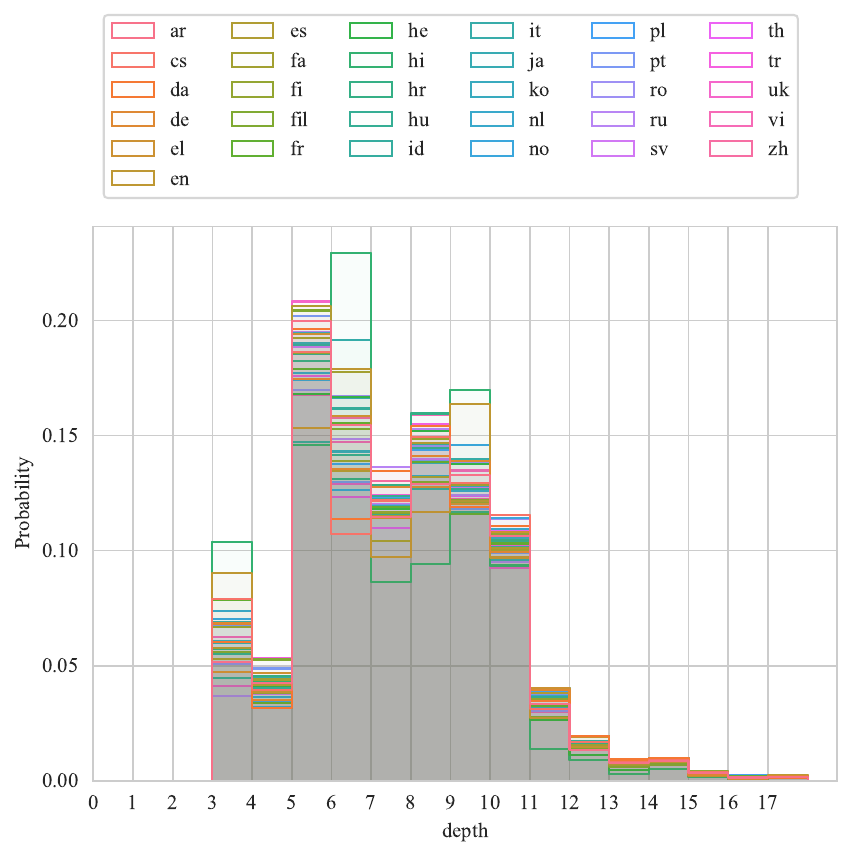}
    \caption{Distribution of depths in the WordNet synset hierarchy across languages.}
    \label{fig:depths_dist}
\end{figure}

\begin{figure*} [t]
    \centering
    
    \begin{subfigure}[t]{\columnwidth}
    \includegraphics[width=\columnwidth]{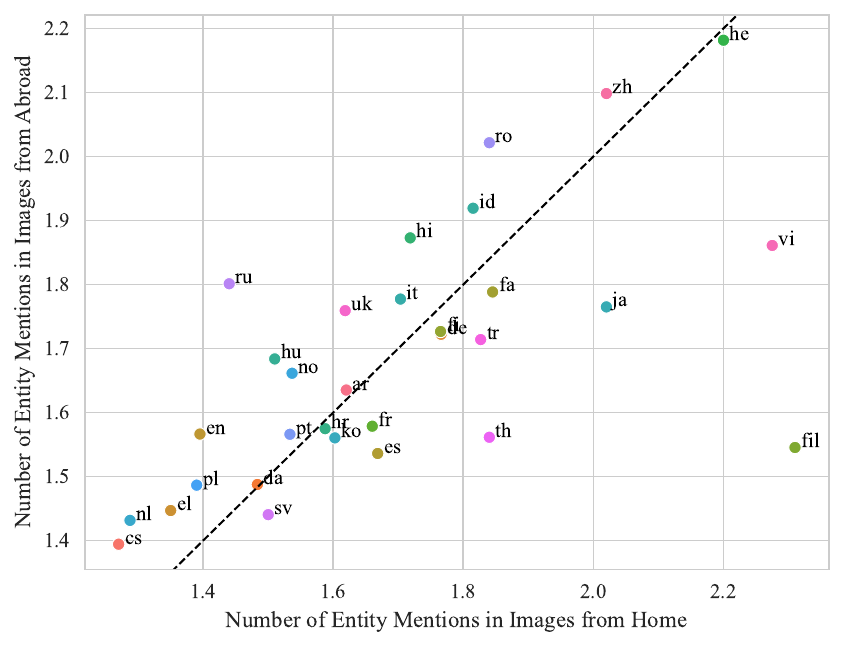}
    \caption{Home vs Abroad}
    \label{fig:object_num_home_abroad}
    \end{subfigure}%
    \begin{subfigure}[t]{\columnwidth}
    \includegraphics[width=\columnwidth]{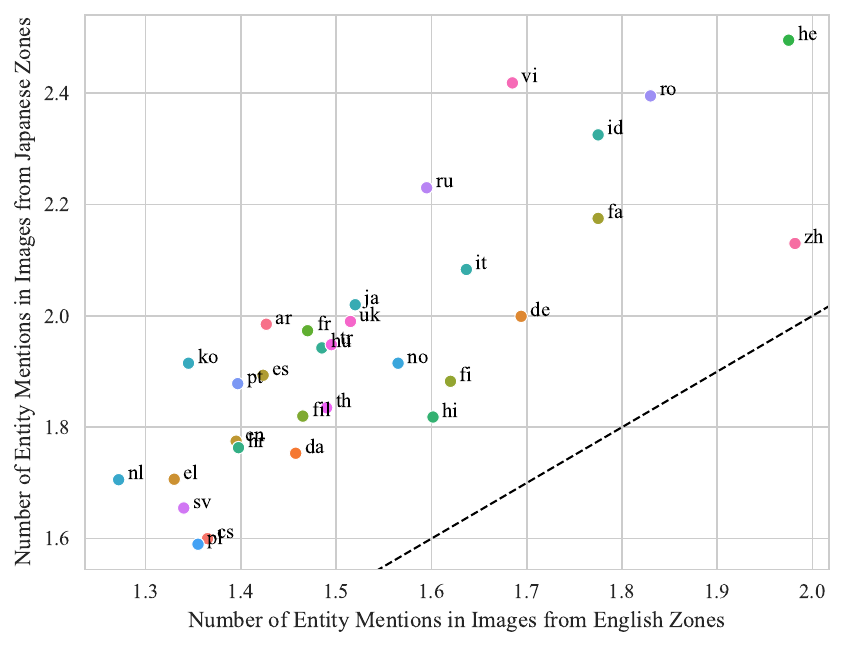}
    \caption{Japanese- vs English-speaking zones}
    \label{fig:object_num_enja}
    \end{subfigure}
    \caption{Average number of entities mentioned by speakers of 31 languages in captions of images captured in different locations. The dashed line indicates identity.}
    \vspace{22pt}
\end{figure*}

\subsection{Granularity} \label{sec:granularity}

Given a hierarchy of categories and their corresponding nominal expressions, from the most generic to the most specific, the level chosen for describing entities is in principle arbitrary. For instance, when describing an image of an armchair, annotators can choose to label it as \textit{furniture}, \textit{chair} or \textit{armchair}, sorted by their depth in the hierarchy in descending order.
\citet{rosch1976basic} introduced the concept of \emph{basic level categories} which are the categories in a hierarchy preferred over more specific subordinates or broader superordinates. If indeed such basic levels exist for a specific language, we would expect the distribution of the levels of mentioned entities to form a two-tailed distribution, with the basic level being used more frequently than the very specific or very general concepts.

We use the depth in the WordNet tree as a proxy for the granularity of an entity category.
Figure~\ref{fig:depths_dist} plots a histogram of synset depths of entities mentioned in each language. Although the number of mentions varies across languages at different levels, all distributions are two-tailed with most mass concentrating in the centre (levels 5--10 in the WordNet hierarchy), in support of the basic-level category hypothesis.

\subsection{Entity Count} \label{sec:object_num}

Does the location of an image affect the number of entities mentioned in its descriptions, due to its familiarity or complexity of the environment? 
To answer this question, we use the location indicators provided for each image in XM3600, where 100 images per language were captured in regions where that language is spoken.
for each language, we first compare the average entity mentions for the 100 images from its relevant regions versus the other 3000 images from abroad.

Figure~\ref{fig:object_num_home_abroad} illustrates that most languages mention around the same number of entities regardless of familiarity, i.e., whether the image was taken at home or abroad. The correlation between these two numbers is strong, with Pearson's $\rho=0.64$, and significant, with $p<10^{-4}$. Major exceptions, however, are found in a group of languages including Filipino, Vietnamese, Japanese, and Thai, which resort to significantly more entities when describing images taken at home. Notably, these languages are spoken in (South) East Asia, suggesting a potential areal effect. %

In addition, we validate one of the studies of \citet{miyamoto2006culture}, namely that patterns of attention are (at least in part) afforded by different environments in images (e.g., Japan vs America). In particular, we compare the average number of entities mentioned in XM3600 pictures captured in Japanese- vs English-speaking zones. Figure~\ref{fig:object_num_enja} shows that Japanese images consistently elicited more entity mentions in all 31 languages (including both Japanese and English), supporting \citet{miyamoto2006culture}'s finding that speakers of both languages count more entities in Japanese images.

\section{Conclusion}
We presented the first large-scale empirical study of how the saliency of entities in image descriptions varies across languages.
Starting from an existing dataset
with culturally diverse images and typologically diverse languages, CrossModal3600, we
developed an automated method to extract entity mentions from multilingual captions and we annotated images with the corresponding entities.

We then conducted a series of experiments on the resulting dataset, finding that:

    \begin{tcolorbox}{%
    \begin{enumerate}
    \item Languages in the same family or area have a mild tendency to mention the same entities, whereas no such effect occurs for featural similarity between languages.
    \item The saliency of some entities is universally high (animate beings), universally low (clothing accessories), or varies significantly across languages (landscape). 
    \item We verify \citet{rosch1976basic}'s theory of basic-level categories, finding that languages universally prefer entities corresponding to synsets in the middle of the WordNet hierarchy (depths 5-10).
    \item The number of entities mentioned is affected by the environment of the image location (e.g., Japan vs Anglosphere), as argued by \citet{miyamoto2006culture}, rather than its familiarity.
    \end{enumerate}
        }%
    \end{tcolorbox}

Our method provides an alternative to small-scale, highly controlled experiments, but is meant to complement them, not replace them. While carefully designed experiments require significant effort in recruiting and training participants, our method is more cost-effective and faster. Therefore, it can serve as an exploratory tool to identify interesting phenomena for later thorough study, and as a large-scale verification for previous findings. We demonstrated both use cases in this work.

\section*{Limitations}
\paragraph{Translation quality.}
Translating captions from all languages into English offers certain advantages (see Section~\ref{sec:caption_translation}). Nevertheless, there are also some drawbacks. One concern is that translation quality may differ across languages, potentially affecting our findings. Additionally, if a specific entity category does not exist in English, it might be translated into a more general category, which could influence the analysis in Section~\ref{sec:granularity}. These issues are especially significant for low-resource languages, leading us to exclude them from the analysis, as mentioned in Section~\ref{sec:experiments}.

\paragraph{Number of annotators per language.}
As mentioned in Section~\ref{sec:en_ja_saliency}, the XM3600 data collection involved only two annotators per language. Consequently, the observed phenomena for a specific language might be significantly influenced by the personal preferences of its annotators. To mitigate this effect, we focused primarily on cross-linguistic patterns observed across all 31 languages. 
In one Section (\ref{sec:en_ja_saliency}), where we did study individual languages, we used an additional, much larger dataset.

\paragraph{Image diversity.}
While examining the XM3600 images, we observed that many images depict the same event. For example, 17 images taken in locations where Ukrainian is spoken (17\% of all Ukrainian images) are from a single event (a large protest), all showing people protesting. This raises the concern that the limited variety in the scenes depicted by XM3600 images might impact our results. Nonetheless, we use XM3600 as it is the only multilingual and culturally diverse image--caption dataset that is currently available.

\section*{Ethics Statement}
We use publicly available resources in our experiments, in accordance with their license agreements. The datasets are fully anonymised and do not contain personal information about the caption annotators or any information that could reveal the identity of the photographed subjects.


\bibliography{anthology,custom}

\clearpage
\appendix

\section{Synset List}
\label{sec:app_synsets}
The list of WordNet synsets used in this study is as follows.

\noindent
\textbf{Root synsets.} 

{\sc
animal.n.01, 

bag.n.01, 

body\_of\_water.n.01,

body\_part.n.01, 

building.n.01,

clothing.n.01,

electronic\_equipment.n.01, 

food.n.01, 

furniture.n.01, 

hand\_tool.n.01, 

home\_appliance.n.01,

jewelry.n.01, 

kitchen\_utensil.n.01,

mountain.n.01, 

musical\_instrument.n.01,

person.n.01, 

plant.n.02,

plaything.n.01,

sky.n.01, 

sun.n.01, 

tableware.n.01, 

timepiece.n.01, 

vehicle.n.01, 

weapon.n.01, 

writing\_implement.n.01
}

\paragraph{Implicit and explicit synsets.}
Please see supplementary materials for the full list of implicit and explicit synsets used in this study.

\section{Synset Extraction} \label{sec:app_extraction}
Here, we elaborate on the synset extraction algorithm. Specifically, we delve into the steps for synset identification and resolving ambiguities.

\subsection{Synset Identification}
To identify synsets, we utilise the WordNet ontology tree but make some modifications to adjust it to image captions:

\begin{itemize}
    \item We remove certain synsets representing meanings not used by annotators in captions. For example, the word \emph{snake} is mapped to the meaning \textsc{snake.n.01} (the animal), but also to the meaning \textsc{snake.n.02} (a treacherous person). As the second meaning is never employed in image captions, we excluded it from the tree.
    \item We unify certain synsets, such as \textsc{food.n.01} (e.g., \textsc{milk.n.01}) and \textsc{food.n.02} (e.g., \textsc{baked\_goods.n.01}) as we would like to consider them as a single category.
    \item The WordNet ontology encodes hyponymy--hypernymy relations between synsets. When synsets are conceptually placed under one root synset but are more visually relevant to another, we relocate them to the latter. For example, \textsc{couple.n.01} (a pair of people) is located under the \textsc{group.n.01} subtree. For our purposes we would like to consider mentions of couples as persons and we therefore move it to the \textsc{person.n.01} subtree.
\end{itemize}

\section{Annotation Guidelines} \label{sec:app_annotation_guidelines}
In this section, we describe the annotation guidelines provided in each of the datasets used in this study.

\paragraph{XM3600.}

The annotation guidelines of XM3600, taken from the original paper, were as follows.

{\fontfamily{Arial}\selectfont
To guide your caption generation, imagine that you are describing the image to a visually impaired friend. The caption should explain the whole image, including all the main objects, activities, and their relationships. The objects should be named as specifically as practical: For example when describing a young boy in a picture, ``young boy'' is preferred over ``young child'', which in turn is preferred over ``person''.

Note: the goal is to generate captions that would be labeled as ``Excellent'' under the Rating guidelines above, but raters should not copy captions from the first phase. We want the raters to generate the captions on their own.

We outline here a procedure that you should try and follow when writing your image caption. Note that not all these steps may be applicable for all images, but they should give you a pretty good idea of how to organize your caption. We will make use of the first image in the table below (the one with the young girl smiling). Note: It is acceptable to make assumptions that are reasonable as long as they don’t contradict the information in the image (e.g.: in the second image below, we use ``families'' in captions 1 and 3 because there seems to be a mix of children and adults though it is not perfectly clear. So it is a reasonable assumption to make and nothing in the image contradicts it. However it is also ok to use ``people''.)

\begin{enumerate}
    \item Identify the most salient objects(s)/person(s) in the image; use the most informative level to refer to something (i.e., ``girl'' rather than ``child'' or ``person''); in the example image: ``girl''
    \item Identify the most salient relation between the main objects; example ``girl standing in front of the whiteboard''
    \item Identify the main activity depicted; in the example image: ``smiling'' as an activity (note that this can also be an attribute of the girl), or ``standing'' as an activity
    \item Identify the most salient attributes of the main object(s)/person(s)/activity(es); in the example image: ``smiling'' and ``young'' as attributes for the girl
    \item Identify the background/context/environment in which the scene is placed; in the example image: ``classroom''
    \item Put everything together from steps 1-5 above; for the example image: ``a smiling girl standing in a classroom'', or ``a young girl smiling in a classroom''.
\end{enumerate}
}

\paragraph{MSCOCO.}
The following instructions were provided to MSCOCO annotators:

{\fontfamily{Arial}\selectfont
\begin{itemize}
    \item Describe all the important parts of the scene.
    \item Do not start the sentences with ``There is''.
    \item Do not describe unimportant details.
    \item Do not describe things that might have happened in the future or past.
    \item Do not describe what a person might say.
    \item Do not give people proper names.
    \item The sentences should contain at least 8 words.
\end{itemize}
}

\paragraph{STAIR-Captions.}
The following instructions were provided to STAIR-Captions annotators:

{\fontfamily{Arial}\selectfont
\begin{enumerate}
    \item A caption must contain more than 15 letters.
    \item A caption must follow the da/dearu style (one of writing styles in Japanese).
    \item A caption must describe only what is happening in an image and the things displayed therein.
    \item A caption must be a single sentence.
    \item A caption must not include emotions or opinions about the image.
\end{enumerate}
}

\end{document}